\documentclass[lettersize,journal]{IEEEtran}
\usepackage{amsmath,amsfonts}
\usepackage{algorithmic}
\usepackage{array}
\usepackage[caption=false,font=normalsize,labelfont=sf,textfont=sf]{subfig}
\usepackage{textcomp}
\usepackage{stfloats}
\usepackage{url}
\usepackage{verbatim}
\usepackage{graphicx}
\usepackage{orcidlink}
\usepackage{booktabs}
\usepackage{multirow}
\usepackage{float}
\usepackage[capitalize]{cleveref}

\def\BibTeX{{\rm B\kern-.05em{\sc i\kern-.025em b}\kern-.08em
    T\kern-.1667em\lower.7ex\hbox{E}\kern-.125emX}}
\usepackage{balance}
\begin{document}
\title{MGE: A Training-Free and Efficient Model Generation and Enhancement Scheme}
\author{Xuan Wang $^{1}$, Zeshan Pang $^{1,*}$, Yuliang Lu $^{1}$, Xuehu Yan $^{1}$\\
\thanks{$*$Equal advising, 1 National University of Defense Technology\\}}

\markboth{MGE: A Training-Free and Efficient Model Generation and Enhancement Scheme}%
{How to Use the IEEEtran \LaTeX \ Templates}

\maketitle

\begin{abstract}
To provide a foundation for the research of deep learning models, the construction of model pool is an essential step. This paper proposes a Training-Free and Efficient Model Generation and Enhancement Scheme (MGE). This scheme primarily considers two aspects during the model generation process: the distribution of model parameters and model performance. 
Experiments result shows that generated models are comparable to models obtained through normal training, and even superior in some cases. 
Moreover, the time consumed in generating models  accounts for only 1\% of the time required for normal model training.
More importantly, with the enhancement of Evolution-MGE, generated models exhibits competitive generalization ability in few-shot tasks. And the behavioral dissimilarity of generated models has the potential of adversarial defense.
\end{abstract}

\begin{IEEEkeywords}
Parameters Distribution, Evolution Algorithm, Auto-Generation, Model Enhancement.
\end{IEEEkeywords}

Deep learning models become the core of various intelligent systems from speech recognition, and image processing to natural language processing. However, challenges exist throughout the life cycle of a deep learning model. At the training stage, the lack of data brings difficulty to model training for specific tasks. The demand for computational power and storage also makes training procedures expensive. At the inference and deployment stage, models face security challenges such as adversarial attacks and backdoor attacks. During the development of a deep learning system, the interpretability of models is closely related to its reliability.

After extensive research and analysis, constructing large-scale model pool can to some extent alleviate the challenges faced by deep learning models at various stages. It holds promising prospects across multiple domains.

\textbf{Few-shot Learning}: The training of traditional deep learning models is often constrained by annotated data and computational resources. Models require a significant amount of data-driven training to achieve optimal performance through iterative processes. However, in reality, data scarcity or even absence is a common challenge, especially in scenarios like few-shot learning\cite{prototypical,relation,MAML,TAML} and zero-shot learning, where the model's generalization is limited by the dataset. Taking the Few-shot problem as an example, it refers to the challenge of training a machine learning model to perform specific tasks with an extremely limited number of training samples. In traditional supervised learning, models typically need a large amount of annotated samples for training, while Few-shot learning aims to overcome the challenge of data scarcity. Such problems often involve dealing with new classes or tasks with extremely limited samples. 

\textbf{Adversarial Learning}: With the emergence of adversarial samples\cite{adv1,CW,FGSM,PGD}, backdoor attacks\cite{BadNets,backdoor_phy,backdoor_trojan}, malicious tampering, and other model attack methods, the vulnerability of deep learning models is gradually exposed, bringing increased attention to model security issues. The fragility of deep learning models is primarily manifested in their sensitivity to adversarial attacks and unknown data. Security research against these attacks requires a large number of models as research subjects for security analysis, such as steganographic analysis methods for models. This method, based on model pool, analyzes the variation of parameters across multiple models and detects potential security threats, thereby enhancing the resistance of models to adversarial attacks and other malicious activities. Simultaneously, model pool can serve as the basis for designing samples for adversarial attacks on models, contributing to the study of adversarial learning and improving the robustness of models, enabling them to better withstand challenges from attackers.

\textbf{Model Similarity and Interpretability}: Currently, Hinton's proposal of neural network similarity representation\cite{HintonSimilarity} confirms that the similarity of models originates from the similarity of network structures, but different network structures can still exist to accomplish the same task. This indirectly suggests that, under the same network architecture, there is similarity in model training patterns, weights, and the positions of activated neurons, which still requires further verification. From facial recognition, speech recognition to intelligent transportation systems, as the application of models in sensitive areas increases, there is a growing demand for the interpretability of models. In fields such as medicine, finance, and law, the decision-making processes of models need to be interpretable and understandable so that decision-makers can better trust and utilize deep learning models. Understanding the internal structure of models is crucial for enhancing model interpretability. All of this is dependent on model pool as support, providing rich data resources for the similarity and interpretability of deep learning models. This helps researchers better understand the behavior and learning patterns of models and engage in innovative algorithm design.

In general, model pool play a crucial role in the fields of deep learning and machine learning, driving the development of algorithms and addressing challenges in practical applications. Acquiring a single deep learning model often requires a substantial amount of data and computational resources through multiple iterations of training, making the construction of large-scale datasets almost an impractical and absurd endeavor. Therefore, for the construction of model pool, it is both a lofty goal and a necessity to explore how to rapidly and automatically generate models tailored to specific tasks without the need for repetitive training.

\begin{figure}[tb]
  \centering
  \includegraphics[height=6.5cm]{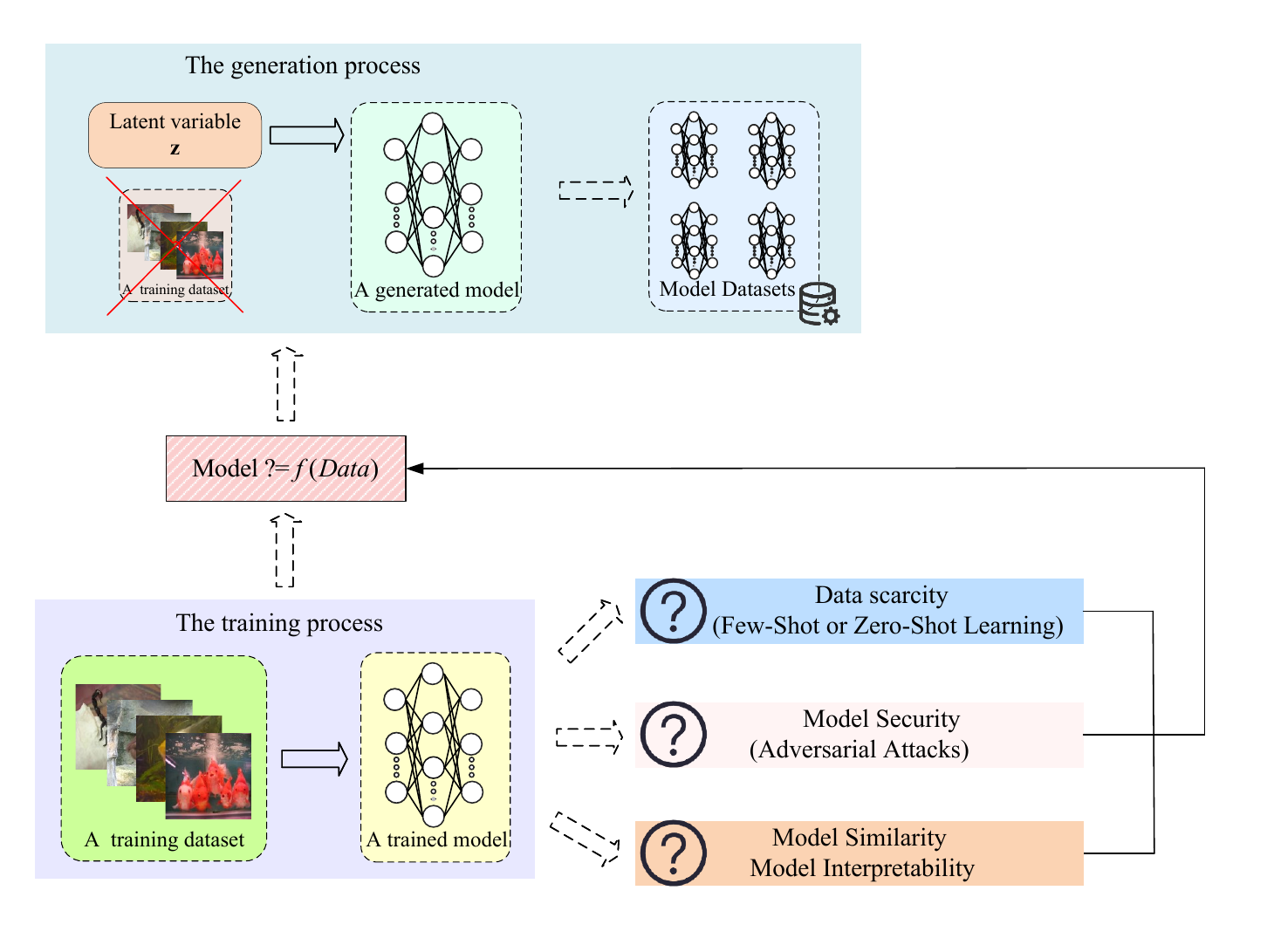}
  \caption{The motivation for model generation.}
  \label{fig1}
\end{figure}

Faced with the strong motivation for the rapid and automatic generation of models, it is essential to analyze that models rely on a large amount of data to drive generation. Factors such as data scarcity, the inherent impact of different datasets on the model's mechanism, and the mathematical reasoning between data and models come into play. Therefore, it is necessary to reduce or even eliminate the impact of data on the model training process. The model generator provides a new approach for this, as illustrated in \cref{fig1}. Without depending on large-scale datasets, and without the need for repetitive training, methods that can rapidly generate model parameters tailored to specific tasks are introduced. The model framework can be specified or randomly selected. This paper proposes a training-free, rapid model generation, and enhancement scheme primarily based on the design of a Generative Adversarial Network (GAN)\cite{gan} architecture. The generator does not require a model training set; instead, it directly generates model parameters for a pre-trained model. These generated parameters are then transmitted to the discriminator. The discriminator assesses the performance of the generated model parameters on a given task and provides feedback to the generator regarding the quality of the generated parameters, assisting in optimizing the generator's output. The discrimination modes within the discriminator may include model accuracy, generalization, robustness, and other criteria.
Considering the vast search space for model parameters and the pseudo-normal distribution characteristics of these parameters, blind searching and generation are not feasible. Therefore, the concept of latent variables is introduced, analogous to the frequency domain space in the model. It sets a well-trained model's high-frequency important parameters as retained, initializes unimportant parameters to 0 as the initial points of latent variables, and uses a complete normal distribution as the gradient convergence direction.
This model generation method can rapidly expand the number of models without requiring a large training dataset and repetitive training. The E-MGE method, while preserving the original advantages of the model, achieves model enhancement by leveraging unimportant parameters. This includes the generalization of the model, applicable to scenarios with small sample sizes, and the robustness of the model, useful for defending against attacks such as adversarial samples. To ensure the convergence direction and speed of performance enhancement, an evolutionary algorithm is introduced. It achieves the rapid acquisition of the optimal model by means of mutation, fusion, and evaluation.
In addition, the fast generation of a large number of model data establishes a solid foundation for studying and analyzing models' safety, providing new ideas for researching the interpretability of models and making the decision-making process more transparent and understandable. The contributions of this paper can be summarized in four points.

\begin{itemize}
\item The model generator directly generates a large number of model seeds, providing a selection of model parameters suitable for specific exclusive tasks. This helps improve the precision and applicability of the models. The approach avoids the resource consumption issues associated with traditional training methods that require large annotated datasets. Through quantitative changes, it achieves qualitative improvements, effectively addressing the problem of model acquisition in scenarios with limited data. On datasets such as MNIST\cite{mnist}, CIFAR-10\cite{CIFAR10}, FashionMNIST\cite{fashionmnist}, and GTSRB\cite{gtsrb}, parameters were generated and tested for dozens of different types of models. For example, on the MNIST dataset, the generated LeNet\cite{lenet} model achieved an average accuracy of 98.25\%, which is nearly 2\% higher than models trained through normal methods.

\item The acquisition of models no longer relies on the accumulation of large amounts of data and repeated training. Combining two dimensions—model parameter distribution and model performance—the parameter space is mapped to the frequency space using latent variables $z$ to generate model parameters for classification. The problem of model generation is transformed into the sampling of latent variables in the potential search space to find the optimal solution. The process of model generation becomes simple and easy to implement.

\item Introducing the heuristic evolutionary algorithm to expedite the generation and enhancement of optimal models for different tasks. This includes the design of operations such as mutation, fusion, and evaluation, showing remarkable results in various scenarios. For instance, in a small sample scenario using the mini-imagenet dataset, the VisionTransformer seeds selected through evolution achieved an accuracy as high as 93.88\%.

\item The model generator can efficiently generate a large number of model samples, making the construction of model pool possible. It can rapidly build various types of model pool under different datasets. For instance, on the FashionMNIST dataset, it only takes 244 minutes to generate 1000 qualified VGG11 models, significantly less than the time required for normal training, accounting for only 1\% of the normal training time. The construction of model pool also serves as the foundation for model security analysis and interpretable model research.

\end{itemize}

\section{Related Works}
In this section, we first review some algorithms that were previously dedicated to filtering important parameters in models. Then, we briefly summarize some evolutionary algorithms on deep neural networks.

\subsection{Evaluation for importance parameters}
n model pruning and compression, the selection of model parameters is often considered, and there are two mainstream approaches: weight-based and activation frequency-based selection.

Weight-based Selection\cite{weight1,weight2}: the main idea is to reduce the number of parameters by setting some weights in the network to zero. This helps in reducing the model's storage requirements, improving inference speed, and sometimes enhancing the model's generalization performance. Common methods include adjusting weight magnitudes and random weight pruning, among others.

Activation Frequency-based Selection\cite{act1,pss}: this method is typically based on the observation that certain neurons in a neural network have a relatively low activation frequency, indicating a potentially smaller contribution to the network's output. These neurons can be pruned to reduce the network's size. The pruning decision is often based on the activation frequency of neurons.

These approaches play a crucial role in model optimization, helping to achieve more efficient and faster neural network models.

\subsection{Evolutionary Algorithms}

In the past, evolutionary algorithms have achieved significant success in a wide range of computational tasks, including modeling, optimization, and design \cite{evomethod}. Inspired by natural evolution, the essence of evolutionary algorithms is to treat possible solutions as individuals in a population. These algorithms generate offspring through mutation and select suitable solutions based on fitness.

Evolutionary algorithms have been introduced to address problems in deep learning. Efforts have been made to minimize human involvement in designing deep algorithms and automatically discover configurations through evolutionary search \cite{LSE,TTevo}, optimizing hyperparameters, and designing deep network architectures. Evolutionary algorithms have demonstrated their capability to optimize deep neural networks \cite{evocnn}. Additionally, \cite{evogan} develops an evolutionary architectural search (EAS) technique to automate the entire design process of the GAN, leading to more stabilized GANs with improved performance. Finally, \cite{LLMevo} introduced a novel evolutionary strategy for optimizing large language models, showing significant potential in solving combinatorial problems.

\section{Methodology}

In this section, we first provide an overview of Model Generation and Enhancement (MGE), including the overall idea, basic definitions, and formulas. Then, we introduce an evolutionary model generation and enhancement algorithm. By elucidating the mutation, fusion, and evaluation mechanisms of E-MGE, we further discuss the advantages of the proposed method. Finally, we summarize the E-MGE training process.

\begin{figure}[!t]
	\centering
	\includegraphics[height=5.5cm]{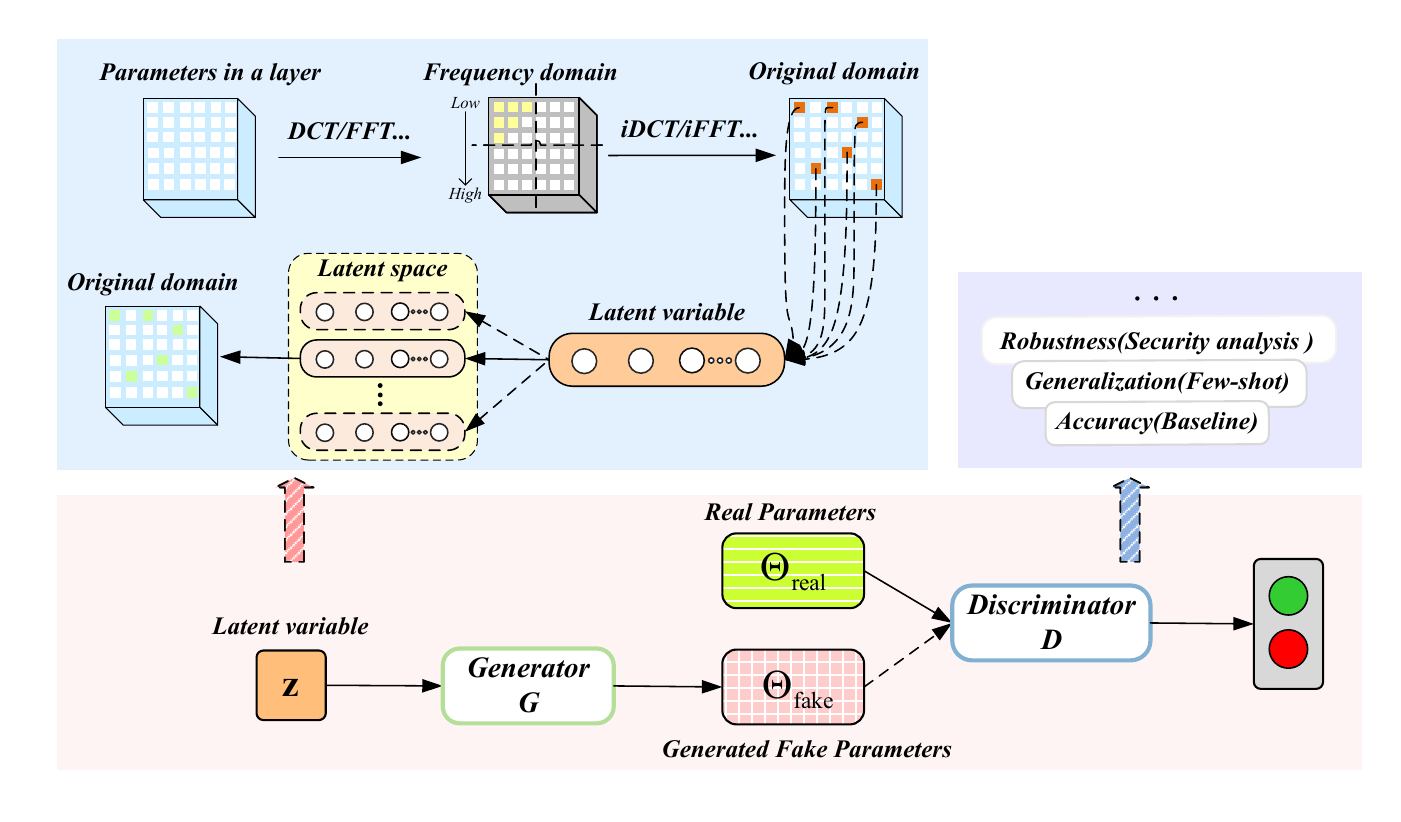}
	\caption{The overview of MGE.}
	\label{generator}
\end{figure}

\subsection{Overview}

The model generation scheme is composed of a generator G and a discriminator D in \cref{generator}.

In the generation process, latent variables of models are generated by mapping model parameters to the frequency domain applying DCT transformation. The latent variables are then fed to the generator G to generate new model parameters. The generator can be adjusted according to given model architecture. 

The quality of generated models are evaluated by the discriminator D, which mainly evaluates the accuracy of model on given tasks. Based on this, additional criteria such as generalization ability and robustness can be included to evaluate the quality of the model according to diverse needs.

The retention of model parameters in the high-frequency part of the frequency domain of a model trained on a specific dataset is aimed at ensuring the accuracy of the model on the selected dataset. The aim is to enhance the generalization and robustness of the model by changing the unimportant parameters in the latent variables, and generate a large number of models that can recognize a specific dataset in a short period of time.

\subsection{MGE}
MGE mainly draws on the adversarial mechanisms of generator $\mathrm{G}$ and discriminator $\mathrm{D}$ in GAN to solve the optimization problem. 
Two characteristics of models are considered in the scheme, the statistical characteristic and the functional characteristic.

The statistical characteristics mainly include the distribution of model parameters. And the functional characteristics mainly include the accuracy, generalization, and robustness of the model.

A pre-trained classification model is first selected as the base model. 
To enhance the generator's generation efficiency while retaining the high-frequency parameters of the energy threshold $t\in[0,1]$, the important parameters are fixed and the generator generates the remaining model parameters.

\subsubsection{Generator}
We propose a parameter generation method based on the model frequency domain, which transforms the model parameters into the frequency domain space, gives a hyperparameter energy threshold $t$, preserves the main energy parameters in the model, and randomly generates unimportant parameters in the $[-z, z]$ interval using a normal distribution generator. The purpose of doing so is to preserve the prototype of the original parameter distribution of the model. Through observation, it is found that most of the model parameters, especially the weight parameters, follow a pseudonormal distribution and are centered around 0. Therefore, from the perspective of model parameter energy, we select important parameters in the frequency domain space to preserve the embryonic distribution of the model, and enhance the utilization of non important parameters in the model to achieve the effect of model enhancement.

On the basis of ensuring the accuracy of the model on the original dataset, enhancing the utilization of unimportant parameters in the model can achieve another form of model multitasking effect. Ideally, it is to compress the efficiency of model neurons and maximize the utilization of model parameters on the basis of a fixed model architecture.

\begin{figure}[h]
\centering
\subfloat[Trained Model]{\includegraphics[width=0.47\linewidth]{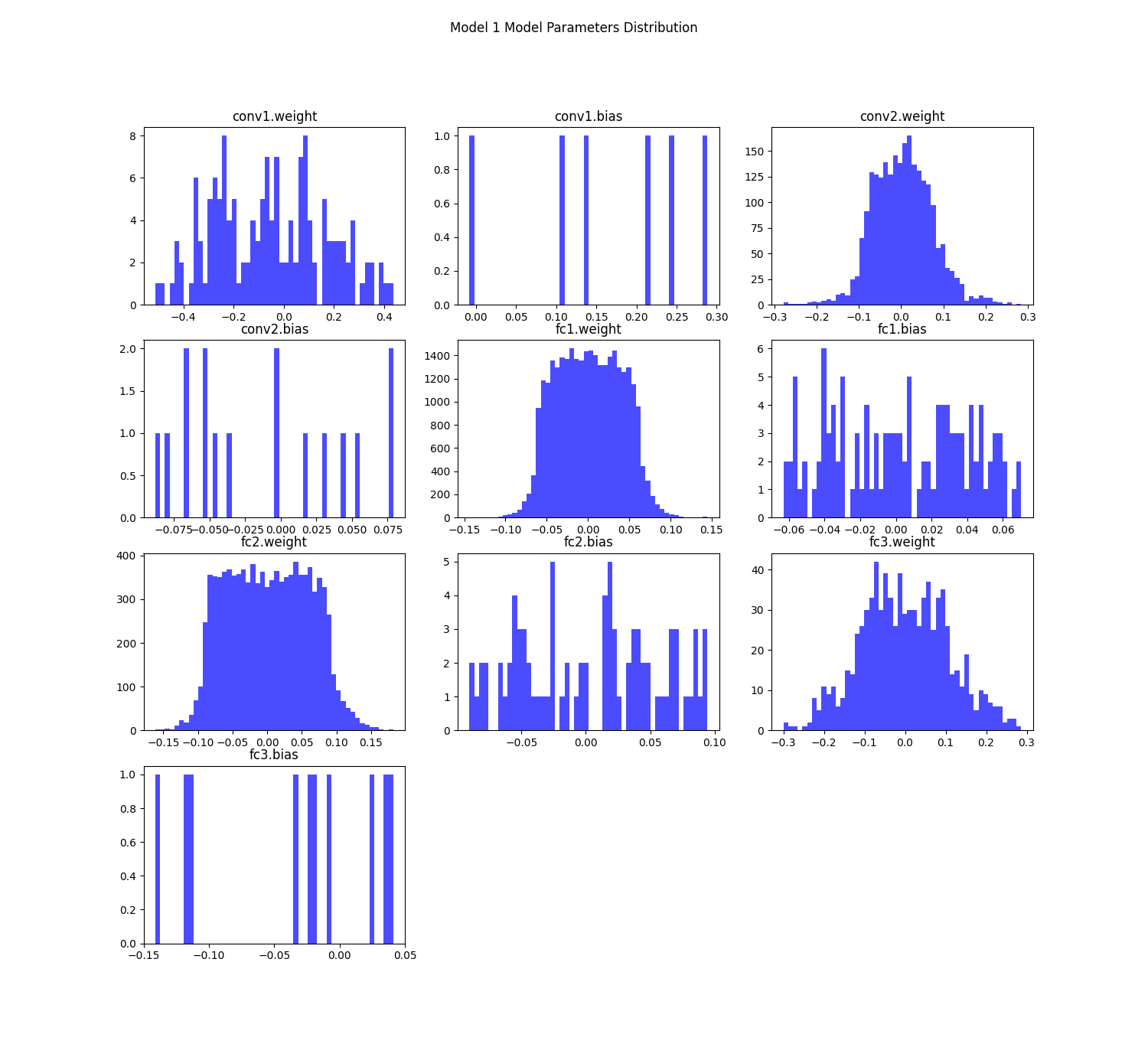}
}
\hfil
\subfloat[Generated Mode]{\includegraphics[width=0.47\linewidth]{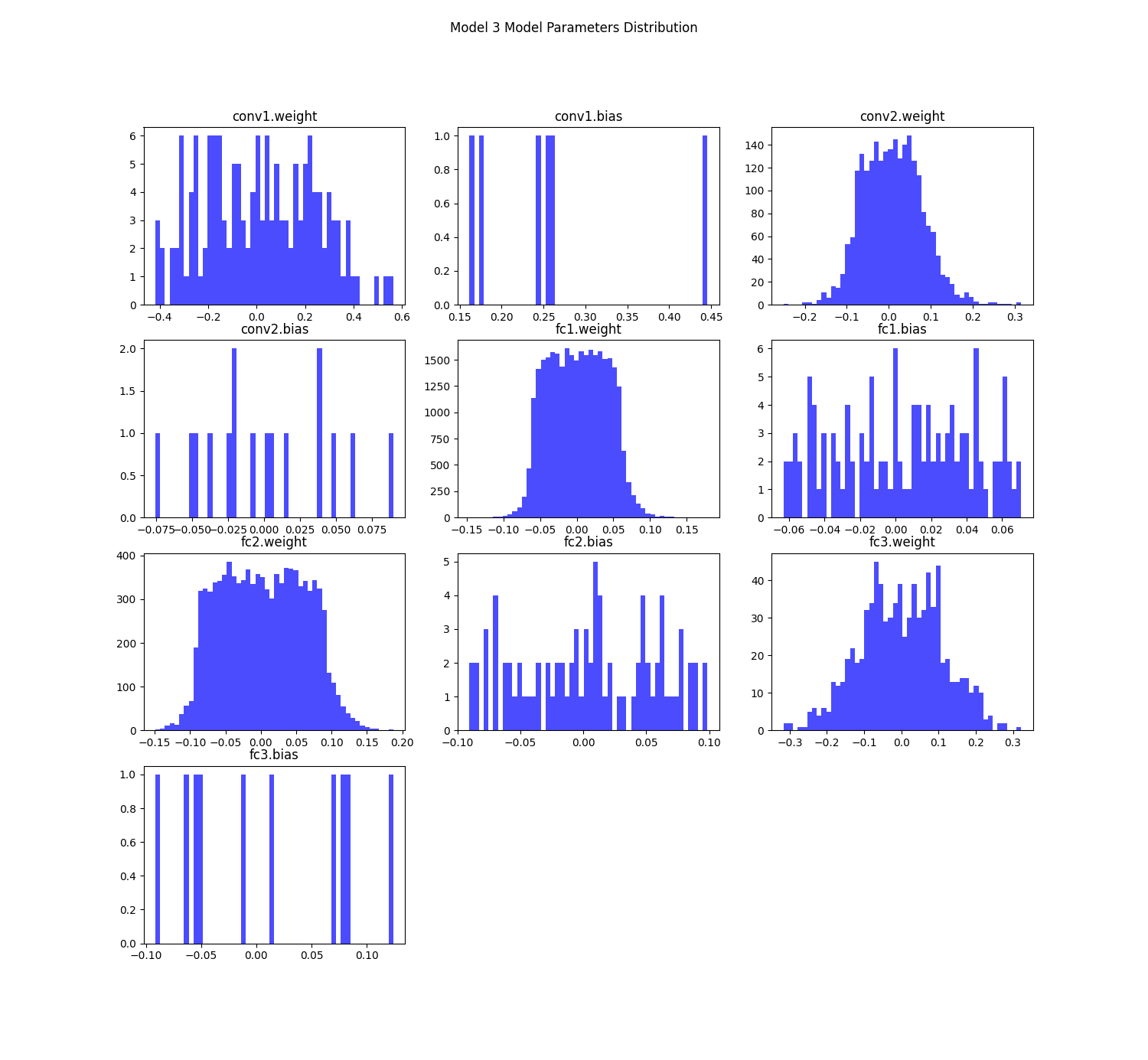}

}

\caption{Distribution of parameters.}
\label{fig:distribution}
\centering
\end{figure}

We present the parameter distribution of the generated model and the parameter distribution image of the trained normal model, as shown in \cref{fig:distribution}. We found that the model parameters generated by the proposed generation method do not change the distribution trend of the model parameters, and only exhibit a surge in individual values. Therefore, from the perspective of model parameter distribution, the model parameters generated by the proposed generation scheme are basically unchanged from those of the normally trained model.

\subsubsection{Discriminator}
In addition to ensuring the statistical characteristics of the model parameter distribution, we add a discriminator to discriminate the accuracy of the model on a specific dataset, with the aim of ensuring that the original functionality of the model is not compromised and consistent with the original functionality of the model obtained from normal training without changing the model distribution. At the same time, enhancing the utilization of unimportant parameters in the model and expanding other characteristics of the model, relevant literature shows that important parameters in existing classification models such as VGG-19 only account for about 8\%. This indicates that the model has a great potential for utilization, so we consider utilizing unimportant parameters to enhance the model and achieve the goal of expanding the excellent model dataset.

MGE mainly draws on the adversarial mechanisms of generator $\mathrm{G}$ 
 and discriminator D in GAN to seek the optimal solution in the game. Firstly, noise samples $z\sim{\mathcal{D}(\mu,\sigma)}$ 
(sampled from a uniform or normal distribution), model layer number $l$, and energy threshold $t$ are collected as inputs. Due to the special nature of the model, we adopt a layered processing method. Generate model parameters $\mathrm{G} (z, l, t)$ for the part of network G output below the energy threshold $t$.

\begin{equation}
\theta^l(z)=\mathcal{N}(-z^l,z^l),z\in[0,0.2]
\end{equation}

Where $z^l$ denotes the latent vector of the $l$-th layer. $\mathcal{N}$ is Normal distribution numerical generator. $\theta^l$ stands for the generated parameters of the $l$-th layer.

\begin{gather}
	\mathrm{G}(z,l,1-t)=\theta^l(z)\
	\\
	\mathrm{G'}(z,l,t)=\theta(t)+\mathrm{G}(z^l,l,1-t)
\end{gather}

The selected parameters of the $l$th layer that are lower than the energy threshold $t$ through frequency domain transformation are generated, while retaining the main energy of the model parameters. After merging, the generated new model parameters are formed.

Meanwhile, utilizing D to distinguish between real model criteria (including accuracy, generalization, robustness, etc.) $f(\theta,x)$ aznd the generated samples $\mathrm{G}(z, l,t)\sim f(\mathrm{G}(z, l,t), x)$.

In MGE, this adversarial training process is described as

\begin{gather}
	\mathop{\min}_{\mathrm{G}}\mathop{\max}_{\mathrm{D}}(E_x(\log(\mathrm{D}(\theta,x))))+E_z\sim \mathcal{D} (\log(1-\mathrm{D}(\mathrm{G'}(z,l,t),x)))
	\\
	\min \mathop{\sum}_{i=1}^{l}|f(\theta,x)-f(\mathrm{G'}(z,l,t),x)|
\end{gather}

The discriminator D

\begin{equation}
	f(\theta,x)=f(\mathrm{G'}(z,l,t),x)+\varepsilon \\ 
\end{equation}

\begin{equation}
      f(\theta,x)<f(\mathrm{G'}(z,l,t),x)
\end{equation}
Where $f$ represents the objective function.

The following is the algorithm of MGE:

\begin{table}[h]
	\begin{tabular}
		{p{0.95\linewidth}} \hline
		\textbf{Algorithm 1.} MGE. Default values $t$=0.8, $n_{D}$=100, $\varepsilon=0.05$, $Freq$=DCT.\\
		\hline
		\textbf{Input}: A trained model $ \mathcal{M}_t$, the number of layer $l$,the $l$-th layer's parameters $\boldsymbol  \Theta^l $. $\mathcal{D}_x$ is the validation set.\par
		\textbf{Output}: A generated model $ \mathcal{M}_g $.\\
		\hline
		{{\small1}:}  \textbf{for} i=0, $...$,$l$ \textbf{do}\par
		{2:}  \qquad  $F_\Theta^i$=$Freq$($\boldsymbol  \Theta^i $ $\leftarrow$ $ \mathcal{M}_t$)\par
		{3:}  \qquad  $\boldsymbol  \Theta^i  $ $\leftarrow$ $iFreq$\{$F_\Theta^i$(high)$\}$||$ $$iFreq$$ $\{ $F_\Theta^i$(low)$\leftarrow$ $\mathrm{G'}(z,l,t)$\}\par
		{4:}  \qquad \textbf{for} k=0, $...$, $n_{D}$ \textbf{do}\par
		{5:}  \qquad \qquad $P^{(i)}$ from a pseudo-normal generator, and $x$ is training data.\par
		{6:}  \qquad \qquad $g_w$ $\leftarrow$ $\nabla_w$ $ \mathcal{D} (\log(1-\mathrm{D}(\mathrm{G'}(z,l,t),x)))$ \{$P^{(i)} \sim (\boldsymbol \Theta^i;z,i)$\}\par
           
		{7:}  \qquad \textbf{end for}\par
           {8:}  \qquad  $ \mathcal{M}_c$ $\leftarrow$ $\boldsymbol  \Theta^i $ \par
           {9:}   \textbf{end for}\par
          {\small10:} \textbf{if} $ f( \mathcal{M}_g, \mathcal{D}_x)>  f( \mathcal{M}_t, \mathcal{D}_x)$ or  $| f( \mathcal{M}_g, \mathcal{D}_x)- f( \mathcal{M}_t, \mathcal{D}_x)|< \varepsilon$ \par
          {11:} \qquad $ \mathcal{M}_g \leftarrow \mathcal{M}_c$ \par
          {12:} \textbf{end if} \\
		 \hline
	\end{tabular}
	\label{tab:Algorithm 1}
\end{table}

In MGE, in order to expand the model dataset, the model is enhanced to improve the utilization of unimportant neurons in the model. The overall idea is to achieve qualitative change through quantitative change. By quickly generating a large amount of model data and testing it under the target task, the optimal solution is heuristic selected through multiple iterations in the search space. Therefore, E-MGE is introduced to evolve GAN, and evolutionary algorithms are introduced in MGE. After generating a large number of models that meet the basic criteria, how to advance to obtain models that meet the additional criteria, while triggering qualitative changes through quantitative changes, and following the principle of survival of the fittest to guide the optimization and updating of the model, accelerating the generation of expected models.

\begin{figure}[tb]
	\centering
	\includegraphics[height=6.5cm]{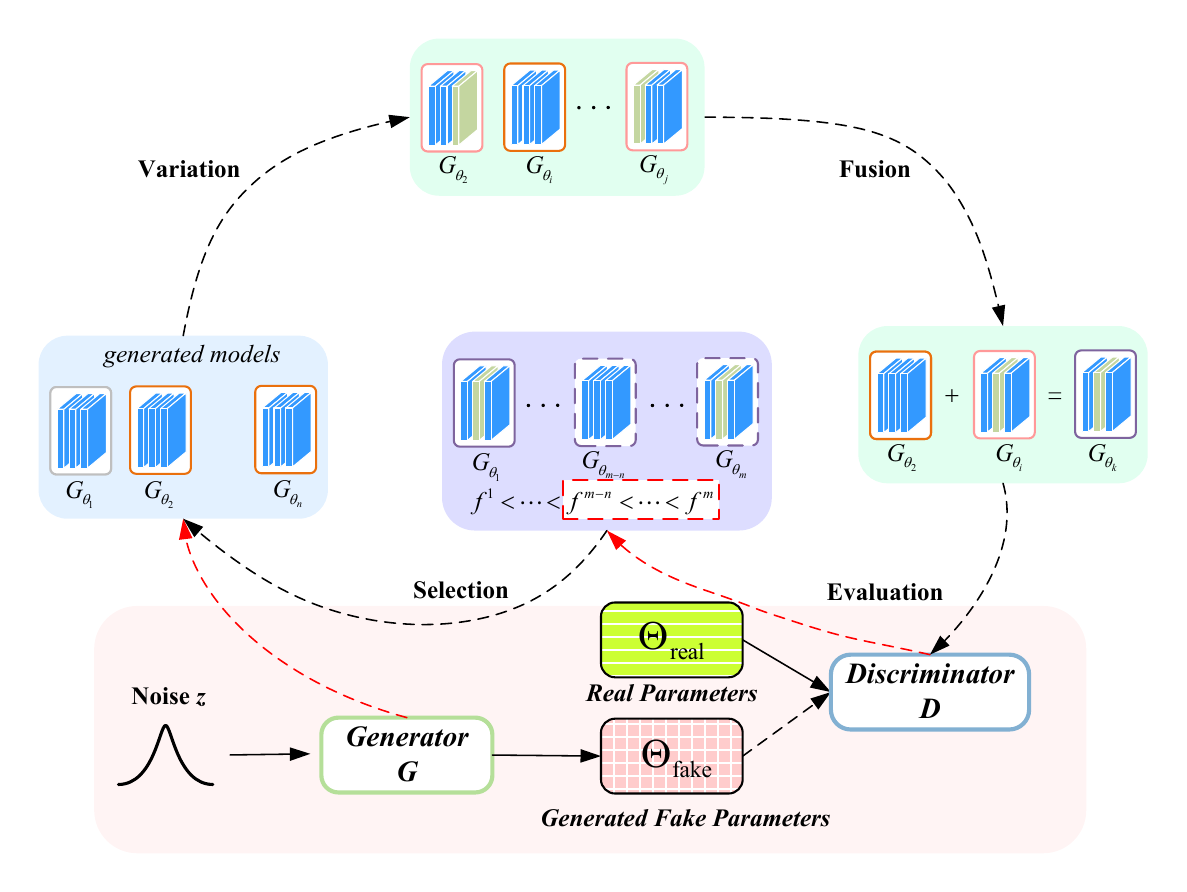}
	\caption{The overview of E-MGE.}
	\label{fig:emodelgan}
\end{figure}

After adjusting and changing the environment (i.e. adding a discriminator $\mathrm{D_{add}}$ with additional criteria), the population of generator(s) {G} is advanced. In this population, each individual represents a possible solution in the parameter space of the generative network G. In order to gradually evolve and adapt to the environment, this means that the evolutionary generator (G) can generate model seeds that are closer to the additional criteria. As shown in \cref{fig:emodelgan}, during the evolutionary process, each step includes three sub stages:

\subsubsection{Mutation}
An individual G in a given population $\mathrm{G}^\theta$, We use mutation operators to generate its descendants $\{\mathrm{G}^\theta_1,\mathrm{G}^\theta_2, \dots \}$. Specifically, multiple copies of each individual or parent are created, and each copy is modified by different mutations. Then, each modified copy is treated as a child copy.

\subsubsection{Evaluation}
For each individual, their performance or quality is evaluated by a fitness function $F(\cdot)$, which depends on the current environment (i.e. discriminator D).

\subsubsection{Fusion}
All children will be selected based on their fitness values, and the worst part will be removed, meaning they will die due to not adapting to the environment. The rest are still alive (i.e., can freely serve as parents) and merge and evolve to the next iteration. Here, with the help of model fusion in federated learning, we mainly use the method of weighted average of model parameters.

In evolutionary algorithms, evaluation refers to the operation of measuring individual quality. To determine the direction of evolution (i.e. individual selection), we designed an evaluation (or fitness) function to measure the performance of evolved individuals (i.e. children). Usually, we focus on the characteristics of two generators: 1) quality and 2) diversity of generated samples. Firstly, we simply input the multiple model seeds generated by the generator into discriminator $D$ and observe the average output, which we name as quality fitness score:

\begin{equation}
	\mathcal{F}_q=\mathbb{E}_z(\mathrm{D_{base}}(\mathrm{G}(z)))
\end{equation}

Please note that during the training process, discriminator D continuously upgrades to the optimal, which reflects the quality of the generator at each evolutionary (or adversarial) step. If the generator obtains a relatively high quality score, the generated samples will deceive the discriminator and the generated distribution will further approach the data distribution.

In addition to generating quality, we also focus on the versatility of generating samples, in addition to the basic functions of generating models such as accuracy. Attention should also be paid to other performance aspects of the model, such as generalization and robustness. Formally, the multifunctionality fitness score is defined as:

\begin{equation}
	\mathcal{F}_d=\mathbb{E}_z(\mathrm{D_{add}}(\mathrm{G}(z)))
\end{equation}

Based on the above two fitness scores, we can ultimately provide an evaluation (or fitness) function for the proposed evolutionary algorithm:

\begin{equation}
	\mathcal{F}=\mathcal{F}_q+\gamma\mathcal{F}_d
\end{equation}

Where $\gamma \geq 0 $ is the trade-off between generation quality and diversity. Overall, a relatively high fitness score F can lead to higher training efficiency and better generation performance.

This section introduced the proposed evolutionary algorithm and corresponding mutation and evaluation criteria, and completed the complete E-MGE training process in Algorithm 1. Overall, in E-MGE, generator {G} is considered an evolutionary population, and discriminator D is an environment. For each evolutionary step, the generator is updated with different targets (or mutations) to adapt to the current environment. According to the principle of survival of the fittest, only individuals who perform well can survive and participate in future adversarial training. Unlike two player games with fixed and static adversarial training objectives in MGE, E-MGE allows algorithms to integrate the advantages of different adversarial objectives and generate the most competitive solutions. Therefore, during the training process, evolutionary algorithms not only ensure the basic functions of the model, but also solve the hidden dangers of unimportant parameters in the model, such as backdoors, and use them to achieve the effect of model enhancement, achieving the goal of completing multiple tasks under limited model resources. Use E-MGE to find better solutions.

\subsection{Experiments}
To evaluate the effectiveness of the generated model and the proposed scheme, we first experimentally observe the impact of selecting important parameters in the frequency domain space on the accuracy of the model. Next, verify the accuracy comparison between the model generated by MGE and the normally trained model, and then increase the generalization and robustness performance testing of the E-MGE model in small sample scenarios and adversarial attacks.

\subsubsection{Settings}
All experiments were trained on Nvidia GTX 1080Ti GPUs. Furthermore, in order to verify the universality and effectiveness of the proposed method through experiments, we chose to test it under different optimizers and learning rates. The experimental settings are shown in \cref{tab:Specific details}.

\begin{table}[h]
	\caption{Specific details of training the model.}
	\centering
\setlength{\tabcolsep}{0.001mm}
	\begin{tabular}{ccccc}
		\toprule
{Model}& {Optimizer} & {Learning rate}	&{Criterion}\\
\midrule
    Resnet-18/VGG16 & SGD  &	0.01 &CrossEntropyLoss \\
  MobileNet-v2/GoogleNet/ResNet& ADAM &0.001 &CrossEntropyLoss\\
   Lenet/Alexnet/VGG11&ADAM &0.001 &CrossEntropyLoss \\
		\bottomrule
	\end{tabular}
	\label{tab:Specific details}
\end{table}

\subsubsection{Parameter selection}
\begin{figure}[h]
\centering
\subfloat[]{\includegraphics[width=0.47\linewidth]{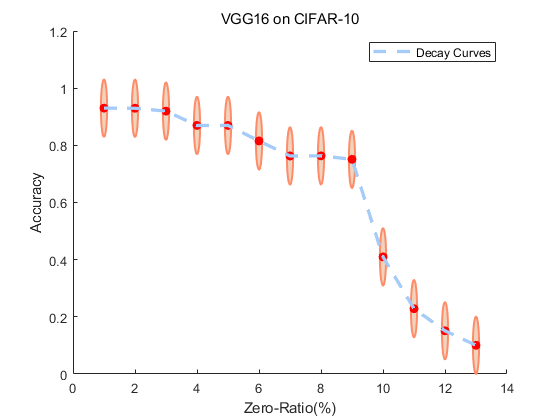}
}
\hfil
\subfloat[]{\includegraphics[width=0.47\linewidth]{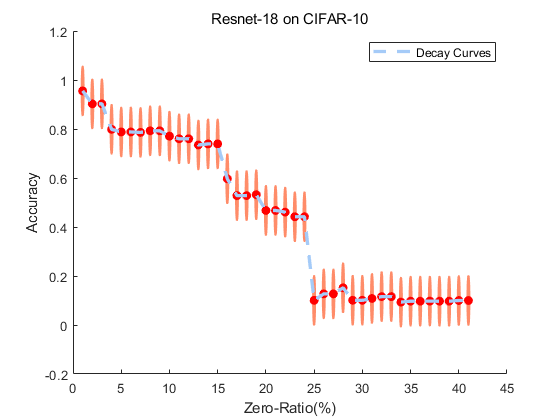}

}
\caption{Decay of model performance after filling.}
\label{fig:decay}
\end{figure}

Regarding the selection of important and unimportant parameters in MGE, after retaining the important parameters, the unimportant parameter positions are uniformly filled with zeros. On the CIFAR-10 dataset, we chose VGG-16 and ResNet-18. When the zero filling ratio was 13\%, VGG-16 had already decreased to the unavailable state of the model, while ResNet-18 rapidly decreased to almost unavailable state at 25\%, which indirectly reflects the availability of model important parameter selection in \cref{fig:decay}.

\begin{figure}[tb]
  \centering
  \includegraphics[height=3.5cm]{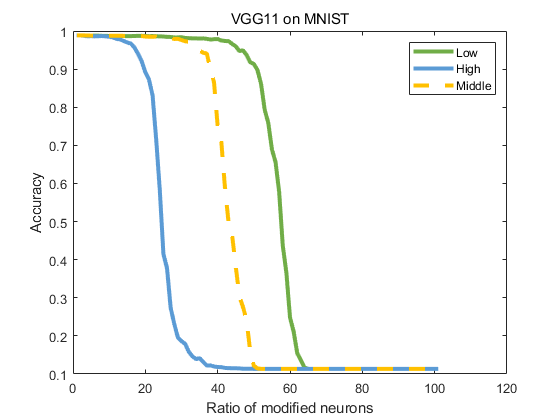}
  \caption{The impact of selecting parameters in different frequency domains on accuracy.}
  \label{fig:changed}
\end{figure}

Considering VGG11 on the MNIST dataset, by modifying the model in the frequency domain with selections of low-frequency, mid-frequency, and high-frequency components, the curve of model accuracy changes as shown in \cref{fig:changed}. We observe that the parameters in the high-frequency region of the model are more sensitive.

We first need to train a LeNet model on the MNIST dataset normally, and perform the following operations on the parameters of each layer of the model. For each layer of model parameters, we need to perform DCT to the frequency domain to obtain the ,odel parameters, and output the frequency domain parameters. 

\begin{figure}[h]
\centering
\subfloat[]{\includegraphics[width=0.47\linewidth]{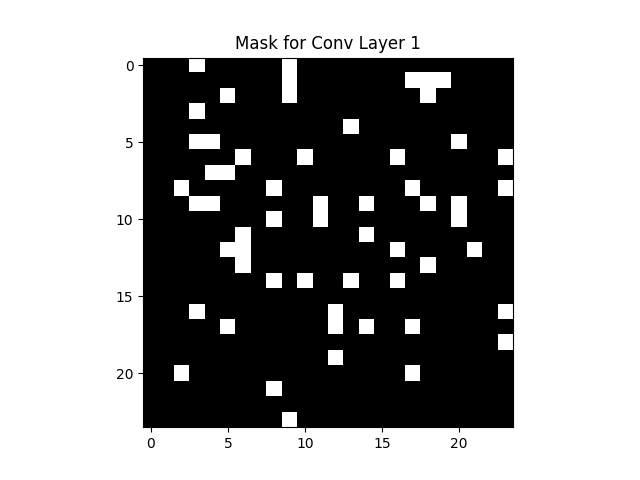}
}
\hfil
\subfloat[]{\includegraphics[width=0.47\linewidth]{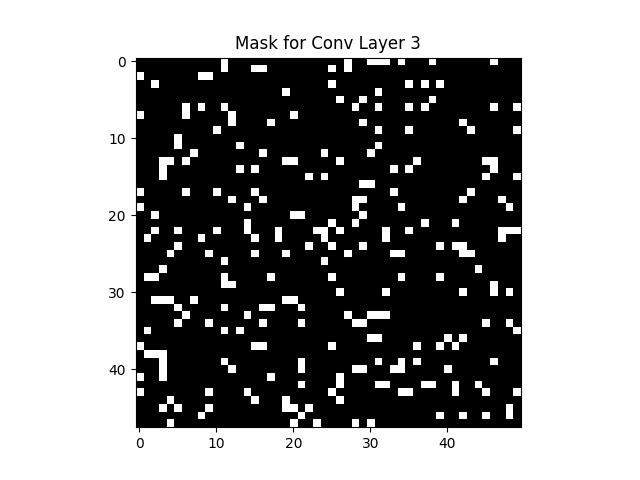}

}
\caption{Mask of unimportant parameters.}
\label{fig:mask}
\end{figure}

In \cref{fig:mask}, 10\% of the mid frequency/low frequency parts of each layer are replaced with a value of 0, and the transformed parameters are returned to the original domain part. After completing the replacement of all layers, the positions of the model parameters that were not replaced by the original training and the parameters with significant changes after replacement in each layer are calculated and recorded as mask positions for unimportant parameters in the model. We present here the unimportant parameters selected by the third convolutional layer in LeNet, with the white position representing the top 10\% of the unimportant parameters selected.

\subsubsection{Baselines}
\begin{figure}[h]
\centering
\subfloat[]{\includegraphics[width=0.47\linewidth]{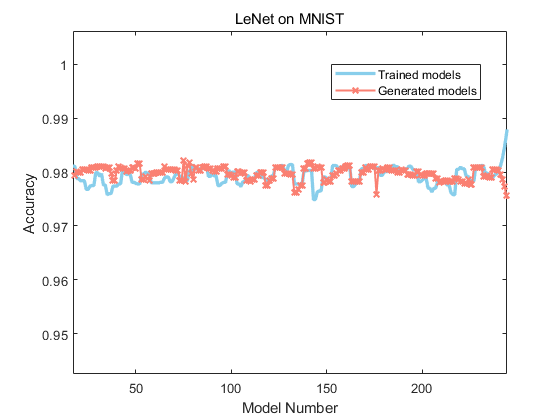}
}
\hfil
\subfloat[]{\includegraphics[width=0.47\linewidth]{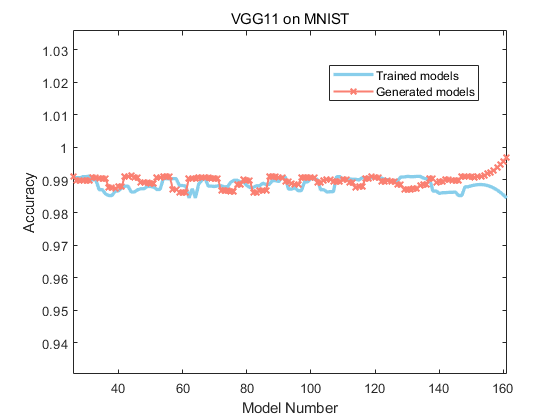}
}
\hfil
\subfloat[]{\includegraphics[width=0.47\linewidth]{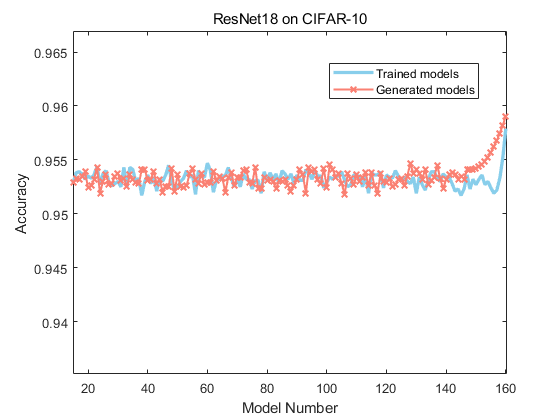}
}
\hfil
\subfloat[]{\includegraphics[width=0.47\linewidth]{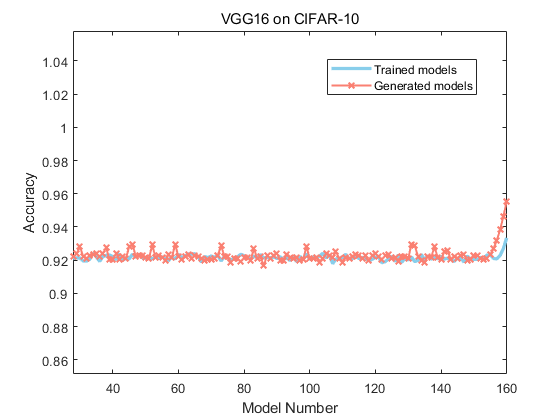}
}
\caption{Accuarcy on Trained models and Generated models.}
\label{fig:train}
\end{figure}

We selected commonly used models LeNet, VGG11, VGG16, and ResNet-18 for accuracy testing on two common datasets, MNIST and CIFAR-10, respectively. The test results are shown in \cref{fig:train}.

According to \cref{fig:train}, the blue line represents the accuracy of the model obtained from normal training, while the red line represents the accuracy of the generated model. It is easy to see that the red and blue lines are intertwined and gradually stabilizing. The side view indicates that the model generator and discriminator have reached the final equilibrium state.

In addition, we tested dozens of different models on four different datasets, and the accuracy of the generated models compared to the trained models is shown in \cref{tab:tgmodelacc}. The results indicate that in terms of classification accuracy, the performance of generated models is almost indistinguishable from that of models trained through normal training, and in some cases, even superior to them.

\begin{table*}[!h]
	\centering
	\caption{Classification accuracy in generated models and trained models.}
	\begin{tabular}{c|c|c|c|c|c|c|c|c}
		\hline
		Dataset & \multicolumn{4}{c|}{MNIST}    & \multicolumn{4}{c}{FashionMNIST} \\
		\hline
		Model & LeNet & AlexNet & VGG11 & ResNet-18 & LeNet & AlexNet & VGG11 & ResNet-18 \\
		\hline
    G-Models& 	98.25\%&	97.92\%&	98.94\% &99.18\% &90.67\% &91.29\% &92.53\% &94.11\%   \\ 
    T-Models& 	98.05\%&	97.50\%&	98.86\% &99.15\% &90.45\% &90.90\% &91.22\% &93.43\%  \\ 
		\hline
		Dataset & \multicolumn{4}{c|}{CIFAR-10} & \multicolumn{4}{c}{GTSRB} \\
		\hline
		Model & LeNet & AlexNet & VGG11 & ResNet-18 & LeNet & AlexNet & VGG11 & ResNet-18 \\
		\hline
    G-Models& 	 	92.13\%&	95.33\%&	98.64\% &98.05\% &96.98\% &98.03\% &97.68\% &98.01\%   \\
    T-Models& 		92.10\%&	95.33\%&	98.24\% &98.08\% &96.70\% &98.02\% &97.64\% &98.01\%   \\ 
		\hline
	\end{tabular}
	\label{tab:tgmodelacc}
\end{table*}

\begin{table*}[!h]
    \caption{Time in generated and trained models.(- represents more than 10w s)}
    \label{tab:tgmodeltime} 
    \begin{center} 
    \setlength{\tabcolsep}{0.1mm}{ 
    \begin{tabular}{c|cccccccc} 
    \hline 
    \multirow{3}{*}{} 
   &\multicolumn{2}{c|}{10}	&\multicolumn{2}{c|}{100}	&\multicolumn{2}{c}{1000}	\\ 

    &\multicolumn{1}{c}{T-Models}	&\multicolumn{1}{c|}{G-Models}	 &\multicolumn{1}{c}{	T-Models}	&\multicolumn{1}{c|}{G-Models}	&\multicolumn{1}{c}{T-Models}	&\multicolumn{1}{c}{G-Models}\\

    \hline
    LeNet(MNIST)& 	744.71s &\multicolumn{1}{c|}{184.57s} & 6688.19s &\multicolumn{1}{c|}{969.40s }&71229.58s &9614.99s \\  \hline
    VGG11(FashionMNIST)& 17569.81s &\multicolumn{1}{c|}{2656.92s} &213824.79s &\multicolumn{1}{c|}{9525.3s }&- &14623.98s \\  \hline
    VGG16(CIFAR-10)&- &\multicolumn{1}{c|}{13224.88s} & - &\multicolumn{1}{c|}{969.40s }&- &9614.99s \\  \hline
    ResNet-18(CIFAR-10)&60624.78s &\multicolumn{1}{c|}{6858.36s} &- &\multicolumn{1}{c|}{9885.18s}&- &35651.72s \\  \hline
    \end{tabular}}
    \end{center}
\end{table*}

In addition to testing the accuracy of the generated models, the time required to generate 10, 100, and 1000 models is also shown in \cref{tab:tgmodeltime}. The average result obtained from 100 iterations is as follows:

\begin{table}[!h]
    \caption{Ratio of time for generated models to trained models.(- represents less than 1\%)}
    \label{tab:tgmodeltimeratio} 
    \begin{center} 
    \setlength{\tabcolsep}{0.1mm}{ 
    \begin{tabular}{c|ccc} 
    \hline 
    \multirow{3}{*}{} 
   &\multicolumn{1}{c|}{10}	&\multicolumn{1}{c|}{100}	&\multicolumn{1}{c}{1000}	\\ 

    &\multicolumn{1}{c|}{$Ratio_{time}$}	&\multicolumn{1}{c|}{$Ratio_{time}$}	 &\multicolumn{1}{c}{	$Ratio_{time}$}	\\

    \hline
    LeNet(MNIST) &\multicolumn{1}{c|}{24.78\%} &\multicolumn{1}{c|}{14.49\%} &\multicolumn{1}{c}{13.50\%} \\  \hline
    VGG11(FashionMNIST) &\multicolumn{1}{c|}{ 15.12\%} &\multicolumn{1}{c|}{4.45\%} &\multicolumn{1}{c}{-} \\  \hline
    VGG16(CIFAR-10)&\multicolumn{1}{c|}{10.94\%}  &\multicolumn{1}{c|}{-}  &\multicolumn{1}{c}{-}\\  \hline
    ResNet-18(CIFAR-10) &\multicolumn{1}{c|}{11.31\%} &\multicolumn{1}{c|}{-} &\multicolumn{1}{c}{-} \\  \hline
    \end{tabular}}
    \end{center}
\end{table}

The time required for generated models is much less than that needed for trained models. The generation time mainly includes the time for the generator to generate unimportant parameters and the time for the discriminator to evaluate the accuracy of the generated models on a specific dataset.

To clearly observe the time saved by the generated models, we defined a metric.

\begin{equation}
	Ratio_{time}=Time_{generated}/ Time_{trained}
\end{equation}

From the \cref{tab:tgmodeltimeratio} species it can be seen that the larger and more complex, the more the number of generated models and the more time saved by the MGE.

\subsubsection{E-MGE}
In order to observe the effect of E-MGE on model enhancement, we selected small sample scenarios and tested the generalization and robustness of the model under adversarial attacks.

We demonstrate the effectiveness and necessity of E-MGE based on MGE and selecting offspring models through evolutionary iteration through experiments. We first test the performance of the LeNet model generated through MGE on the MNIST dataset under C\&W attack\cite{CW}.

\subsubsection{Behavioral similarity}
The behavioral similarity of generated models are difficult to evaluate since the model parameters are hard to interpret. We instead demonstrate the dissimilarity by conducting adversarial attacks on the generated models. Specifically, we implement C\&W attack\cite{CW} on base model and obtain 100 adversarial examples. These adversarial examples are then tested on 52 generated model.

\begin{figure*}[tb]
	\centering
	\includegraphics*[width=1\textwidth]{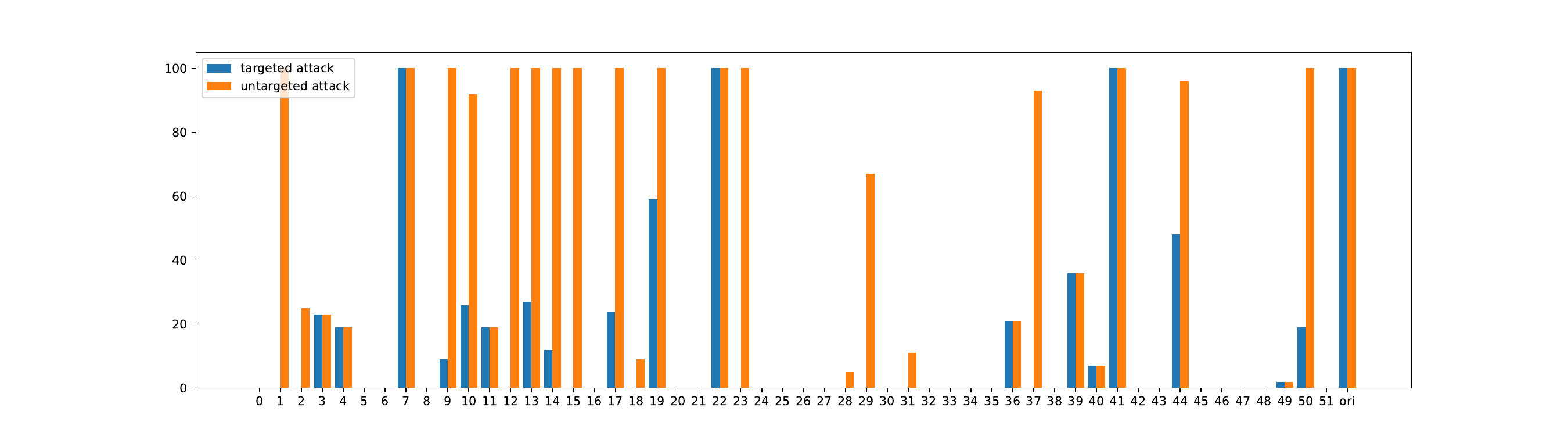}
	\caption{Performance of generated models under C\&W attack.}
	\label{adv}
\end{figure*}

\begin{table}[h]
	\caption{Comparison of model robustness under fgsm attack with different perturbations.}
	\centering
\setlength{\tabcolsep}{0.5mm}
	\begin{tabular}{c|c|c}
		\toprule
{Perturbation}	& {Original Model} & {Evolution Model}\\
\midrule
    epsilon=0.01 & 81.37\%  &93.44\% \\
    epsilon=0.1& 80.39\% &93.12\% \\
    epsilon=1&76.89\% &92.58\% \\
		\bottomrule
	\end{tabular}
	\label{tab:diffper}
\end{table}
The attack result is shown in \cref{adv}. The labels on horizontal axis are the numbers of generated models and 'ori' is the base model. The bars of "targeted attack" refer to the cases where adversarial examples are misclassified to the target category we set when generating adversarial examples. And the "untargeted attack" bars indicate the cases where adversarial examples are misclassified to any categories other than the original category.

The result suggest that the generated models do not necessarily inherit the vulnerabilities of the base model and behave dissimilar to the base models in adversarial scenarios. This characteristic may be further used in the defense against adversarial attacks.

\subsubsection{Robustness}
Through the behavioral similarity experiments described above, we find that the partially generated models are highly robust. On the side, it can be seen that the generated models have a certain degree of diversity, providing a good foundation for the next step of evolution.

We set the number of iterations to $N$, the number of parents to $n=10$, and the number of mutations to $j=10$. There are a total of $n+j$ models available for selection for fusion, with a fusion quantity of $m=20$. The top 10 models with excellent performance are chosen as parents for the next iteration. The hyperparameter for the evaluation function is denoted as $\gamma$.

Next, we choose FGSM attack \cite{FGSM} to add different perturbations (epsilon) to the image dataset and observe the effectiveness of the model generated by E-MGE. We conducted $N=100$ to select models that performed well in robustness testing (fitness function) for perturbed samples for evolution.

According to the results in \cref{tab:diffper}, we can clearly see that the final filtered evolutionary model has strong robustness and can accurately recognize the noisy images.

\subsubsection{generalization}

We test the accuracy of the generated model on mini-ImageNet dataset and the transferability of the model itself. It is important to emphasize that simply providing model seeds that are likely to perform well on a particular task during the evolutionary process is disjointed from continuing to train model seeds in the next step.

\begin{figure}[h]
\centering
\subfloat[]{\includegraphics[width=0.47\linewidth]{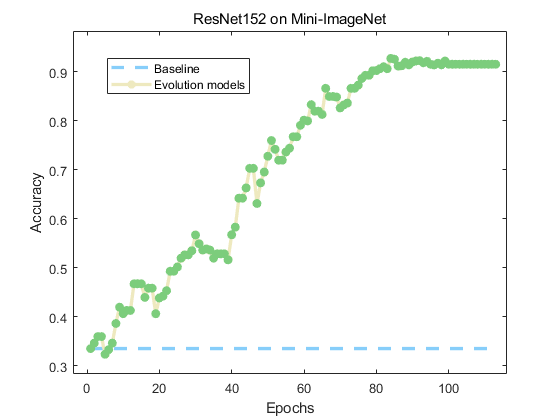}
}
\hfil
\subfloat[]{\includegraphics[width=0.47\linewidth]{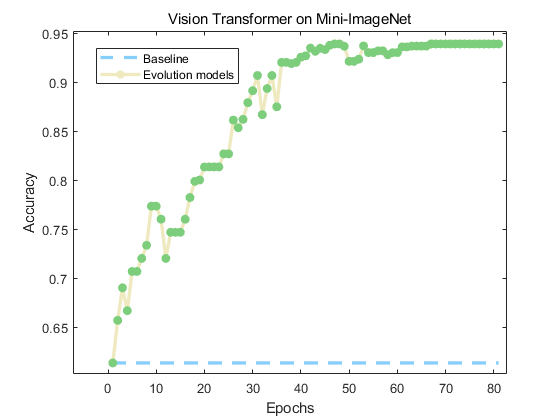}

}
\caption{The accuracy during evolution. ($N=150$)}
\label{evolution}
\end{figure}

\cref{evolution} have shown that E-MGE has the ability to strongly screen model seeds. 

The model seeds (similar to pre-trained models) screened by the E-MGE method on Vision Transformer can reach 93.88\% accuracy after fine-tuned training.

\subsection{Discussion}
Through experiments, it was found that there is redundancy in the parameter space in most models, which can be utilized. We only selected 10\% of unimportant parameters for model enhancement, which enhances the generalization or robustness of the model and achieves excellent results. This indirectly indicates that different models have different contributions to the transfer effect. In fact, we have found that there may be perfect models in the model, that is, a certain model has exceptionally excellent performance in a specific task, and can quickly converge and achieve good results through fine-tuning. E-MGE is a method for finding perfect seeds.

The limitation of the proposed solution is that, regarding the model dataset, all we can do is to perform certain data augmentation on the original model data, such as image steganography and image enhancement, which have a certain expansion effect on the original model samples. However, we still need to consider studying and analyzing the diversity of the model dataset. The next step of the work will be to expand the model dataset based on the existing method of constructing the model dataset, while introducing a diffusion model. By re learning the noise model and directly generating models related to the data, the goal of expanding the diversity of model samples will be achieved.

\subsection{Conclusion}
In this article, we propose MGE to automatically generate a large number of DNN models without the need for training and efficiently generate and enhance models, driven by the construction of a model dataset. In order to achieve rapid enhancement of the model, E-MGE is proposed, and evolutionary algorithms are introduced to efficiently screen out model seeds that perform well under the fitness function through mutation, fusion, and evaluation. Experiments have shown that MGE can generate a large number of models in a short period of time with good model generation effects, while E-MGE can effectively improve model performance and enhance the model. Future work will focus on further exploring the relationship between model diversity and model parameters, and further improving the integrity of the constructed model dataset.

{
\bibliographystyle{ieee_fullname}
\bibliography{aaa}
}

\end{document}